\pdfoutput=1

\documentclass[11pt]{article}

\usepackage[final]{acl}

\usepackage{times}
\usepackage{latexsym}
\usepackage[T1]{fontenc}
\usepackage[utf8]{inputenc}
\usepackage{comment}
\usepackage{microtype}
\usepackage{inconsolata}
\usepackage{graphicx}
\usepackage{url}
\usepackage{tablefootnote}
\usepackage{booktabs}
\usepackage{efbox,graphicx}
\efboxsetup{linecolor=black,linewidth=0.3pt}
\usepackage{multirow}
\usepackage{multicol}
\usepackage{setspace}
\usepackage{xcolor}

\title{\textsc{MetaphorShare}: A Dynamic Collaborative Repository \\ of Open Metaphor Datasets}

\author{Joanne Boisson, Arif Mehmood \and
Jose Camacho-Collados \\
    Cardiff NLP, School of Computer Science and Informatics\\ 
  Cardiff University, United Kingdom\\
  \texttt{\{boissonjc,mehmooda3,camachocolladosj\}@cardiff.ac.uk} \\}

\begin{document}
\maketitle
\begin{abstract}

The metaphor studies community has developed numerous valuable labelled corpora in various languages over the years. Many of these resources are not only unknown to the NLP community, but are also often not easily shared among the researchers. Both in human sciences and in NLP, researchers could benefit from a centralised database of labelled resources, easily accessible and unified under an identical format. To facilitate this, we present \textsc{MetaphorShare}, a website to integrate metaphor datasets making them open and accessible. With this effort, our aim is to encourage researchers to share and upload more datasets in any language in order to facilitate 
metaphor studies and the development of future metaphor processing NLP systems. The website has four main functionalities: upload, download, search and label metaphor datasets. It is accessible at \url{www.metaphorshare.com}.

\end{abstract}

\section{Introduction}

The topic of figurative language processing has been addressed since early years of artificial intelligence research \cite{Martin:1990,Fass1997ProcessingMA, Kintsch2000}, inspired by philosophical \cite{richards1936philosophy}, linguistic \cite{wilks-pref} and then cognitive science theories \cite{LakoffJohnson80} and experimental studies \cite{katz-norms}. In spite of the influence of the different disciplines on one another, the resources created to train and evaluate NLP models have often been distinct from the resources created for metaphor studies in other fields, with the notable exception of Master Metaphor List \cite{lakoff1994master} and the VU Amsterdam Metaphor Corpus \cite{Steen:2010aa}.

In recent years, metaphor processing has attracted more and more attention with the progress made possible by transformer-based language models, for example \citet{mao-etal-2019-end} or \citet{ zeng-bhat-2021-idiomatic}, and in particular the large LMs such as GPT-3 \cite{GPT3} used by \citet{wachowiak-gromann-2023-gpt} for metaphor identification. Several dedicated workshops with shared tasks \cite{flp-2022-figurative, fig-lang-2024-figurative,sharma-etal-2020-semeval, tayyar-madabushi-etal-2022-semeval} have been organized, leading to the creation of more resources.
The proper handling of figurative language by models is of crucial importance for improving performance in downstream tasks \cite{han-etal-2022-hierarchical, li2024findingchallengingmetaphorsconfuse}.

\begin{figure*}[t!]
\efbox{\includegraphics[width=.983\textwidth]{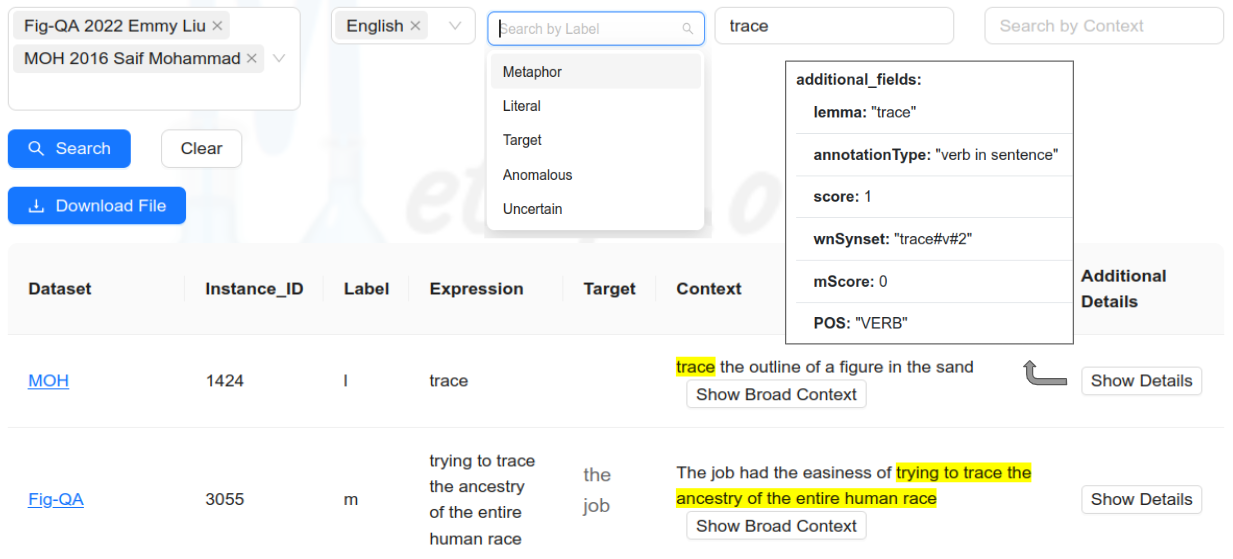}}
\caption{\textsc{MetaphorShare} search page. Specific datasets, languages, and tag types can be selected, and a text-based search within tagged expressions or into the entire text is implemented. Additional features provided with the record appear when clicking the \textit{Show Details} button.}
\label{fig:search}
\end{figure*}

From the perspective of Humanities, tools for searching corpora such as Sketch Engine \cite{kilgarriff2014sketch}, annotation tools \cite{semantic_anno_2008} and data analysis softwares have become part of the standard corpus-based research methodology. Effective adaptive support for metaphor identification and interpretation is the next step of this on-going inter-disciplinary collaboration. There were two limitations until recently: metaphor identification systems were not accurate enough to be used on free text, and different researchers are looking for different metaphors.

On one hand, numerous datasets labelled with metaphoric usage of words are created by researchers in many languages every year, and remain out of the scope of NLP research, tool development, and evaluations of the systems. This happens because most datasets are published in non NLP venues, they might not be publicly available, or might be tagged in an under-specified format that does not allow direct automatic processing (e.g. tagging precisely a metaphoric expression within a text). On the other hand, metaphor studies researchers often work on domain specific corpora, languages other than English - for which NLP has developed the most resources - and use different definitions or retain different types of metaphors for different projects, with no convenient way to systematically compare their labels with other researchers, or to rely on an existing metaphor automatic labelling tool that answers their project specific need. 

We propose to help speeding up collaborations between AI/NLP communities and linguistics/metaphor studies by facilitating the unification of dataset formats and the access to existing resources by everyone, while preserving the information encoded in the original datasets. MetaphorShare is a website designed at sharing new and publicly available existing resources (see Figure \ref{fig:search}). It is an online dynamic repository where anyone can upload or download open metaphor datasets. Our online labeling tool also offers the possibility to directly create a dataset, and optionally share it. Our hope is that the website will create a new synergy between different components of metaphor studies, limit the loss of datasets, catalog them, and facilitate their automatic processing through a unified format. Our target audience comprises anyone conducting corpus-based research on metaphors, including NLP researchers for the development of metaphor processing and identification systems\footnote{A short introduction to \textsc{MetaphorShare} is available at \url{https://youtu.be/Fi48SOjueEE}}.
 
Recent advances in metaphor processing are presented further in Section \ref{sec:related_work}. The diversity of existing resources and the unified dataset input format is presented Section \ref{sec:formats}. 
The website is structured around four main functionalities that are described in Section \ref{sec:metaphorshare}: upload,
download, search and label metaphor datasets.
Section \ref{sec:evaluation} shows the potential of our framework for facilitating NLP experiments, by evaluating a RoBERTa \cite{ROBERTA} model on a cross-dataset classification task.

\section{Related work}
\label{sec:related_work}

\paragraph{Who studies and labels metaphors?} 

\citet{cameron2010metaphor} summarizes the landscape of metaphor studies. The core role played by \citet{LakoffJohnson80}'s Conceptual Metaphor Theory (CMT) in Cognitive Science had an impact on the methodology developed for the analysis of metaphors in Linguistics and Literature. Metaphor analysis also became a standard approach in Anthropology, Educational Research, Political Science, or Management Research. The purpose of metaphor analysis in these disciplines is to uncover latent meaning present in discourse on a studied topic, that is sometimes conveyed elusively in corpora. Recently, \citet{baleato-rodriguez-etal-2023-paper} introduced propaganda modeling, integrating metaphor identification to the automatic detection of persuasive intentions.

\paragraph{Metaphor related websites.} \citet{veale-li-2012-specifying} develop a system of metaphor interpretation and generation, \textsc{Metaphor Magnet}, that relies on the harvest of stereotypes from Google n-grams. The stereotypical attributes associated with concepts are leveraged for the suggestion of relevant metaphors and their interpretation (e.g. given \textit{Google is –Microsoft}, the system outputs \textit{giant} with properties like \textit{lumbering} and \textit{sprawling}). More recently, \citet{mao-etal-2023-metapro} release an end-to-end domain independent metaphor identification and interpretation website for free text: \textsc{MetaPro} tags metaphoric tokens or multi-word expressions in sentences, generates  paraphrases, and outputs a more abstract concept mapping derived from the analysis of the sentence. The authors list limitations due to the lack of training data for inference and extended metaphors. The main difference with these two initiatives and ours is that our aim is to integrate already labeled datasets into a unified repository, encouraging the structuring  of resources and expression of needs from the metaphor study community, before building automatic annotation tools. Similarly to ours, MetaPro also uses several metaphor datasets for training the models \cite{mao-etal-2019-end}, but in this case they do not mention more than six datasets, all of them often used by the NLP community.

\paragraph{Annotation tools.} Many open source
(e.g. Label Studio, Dataturks, Doccano \cite{doccano} and
Potato \cite{pei-etal-2022-potato}) and proprietary
(e.g. LabelBox, Prodigy or Amazon Mechanical Turk)
annotation tools exist. Our purpose is not to compete with these services but rather to offer a simple targeted experience for metaphor annotation, in which the output is directly mapped to the unified repository format. This way, non-expert users can avoid the need for learning a general tool, as well as having to learn how to modify given labeled outputs, which would not be trivial for most.

\paragraph{Metaphor processing in NLP.} One important usage of our website is to facilitate the creation of personalised metaphor processing models. Metaphor processing in NLP comprises many methods developed for metaphor identification \cite{turney-etal-2011-literal,tsvetkov-etal-2014-metaphor,mao-etal-2019-end,wachowiak-gromann-2023-gpt}, but also generation \cite{veale-2016-round,stowe-etal-2021-metaphor,chakrabarty-etal-2021-mermaid}, textual \cite{mao-etal-2018-word} and multimodal \cite{kulkarni-etal-2024-report} interpretation, metaphor understanding through entailment \cite{agerri-etal-2008-textual,chakrabarty-etal-2021-figurative,stowe-etal-2022-impli}, among other tasks. \citet{Ge2023} provide a comprehensive recent survey on the topic.

\begin{table}[]
    \centering
    \footnotesize
    \begin{tabular}{l@{\hspace{2pt}}l@{\hspace{2pt}}l@{\hspace{0pt}}r}
    \toprule
    \textbf{Name}&
    \textbf{Reference} &
    \textbf{N}&
    \textbf{\%M}\\\midrule
         \multicolumn{4}{l}{\textbf{Words in syntactically constrained sentences }}  \\
         \multicolumn{4}{l}{\textbf{(Psycholinguistics)}}\\\midrule
         JANK&\citet{Jankowiak2020}&240&50\\
         CARD\_V&\citet{Cardillo-2010}&280&50\\
          CARD\_N&\citet{Cardillo-2010,Cardillo2017}&512&50\\\midrule
          \multicolumn{4}{l}
          {\textbf{Words in natural short contexts (NLP)}}\\\midrule
         MOH&\citet{mohammad-etal-2016-metaphor}&1632&25\\
        NewsMet&\citet{joseph-etal-2023-newsmet}&1205&49\\
     TSV\_A & \citet{tsvetkov-etal-2014-metaphor}&1945&50\\
        GUT&\citet{gutierrez-etal-2016-literal}&8591&54\\\midrule
        \multicolumn{4}{l}{\textbf{MWE in natural long context (NLP)}}  \\\midrule
        PVC& \citet{TuRoth12}&1348&65\\
        MAD& \citet{tayyarmadabushi-etal-2022-semeval}&4558&48\\
        MAGPIE& \citet{haagsma-etal-2020-magpie}&48395&75\\\midrule 
        \multicolumn{4}{l}
        {\textbf{Words sampled from VUAC (MIPVU)}}\\\midrule
        VUAC\_BO  &\citet{boisson-etal-2023-construction}&39223&52\\
        TONG&\citet{tong2024metaphor}&1428&46\\
    \bottomrule
    \end{tabular}
    \caption{Twelve metaphor datasets included in the \textsc{MetaphorShare} repository, with a single metaphor/literal tag per entry. Short contexts consist of one word or a short sentence. Long contexts correspond to multiple sentences. Datasets derived from VUAC have a document level context. The N column shows the number of instances in the datasets. \%M indicates the percentage of expressions labelled as metaphorical.}
    \label{tab:ten_datasets}
\end{table}

\begin{figure*}[t!]
\begin{center}
\efbox{
\includegraphics[width=.82\textwidth, height=7.8cm]{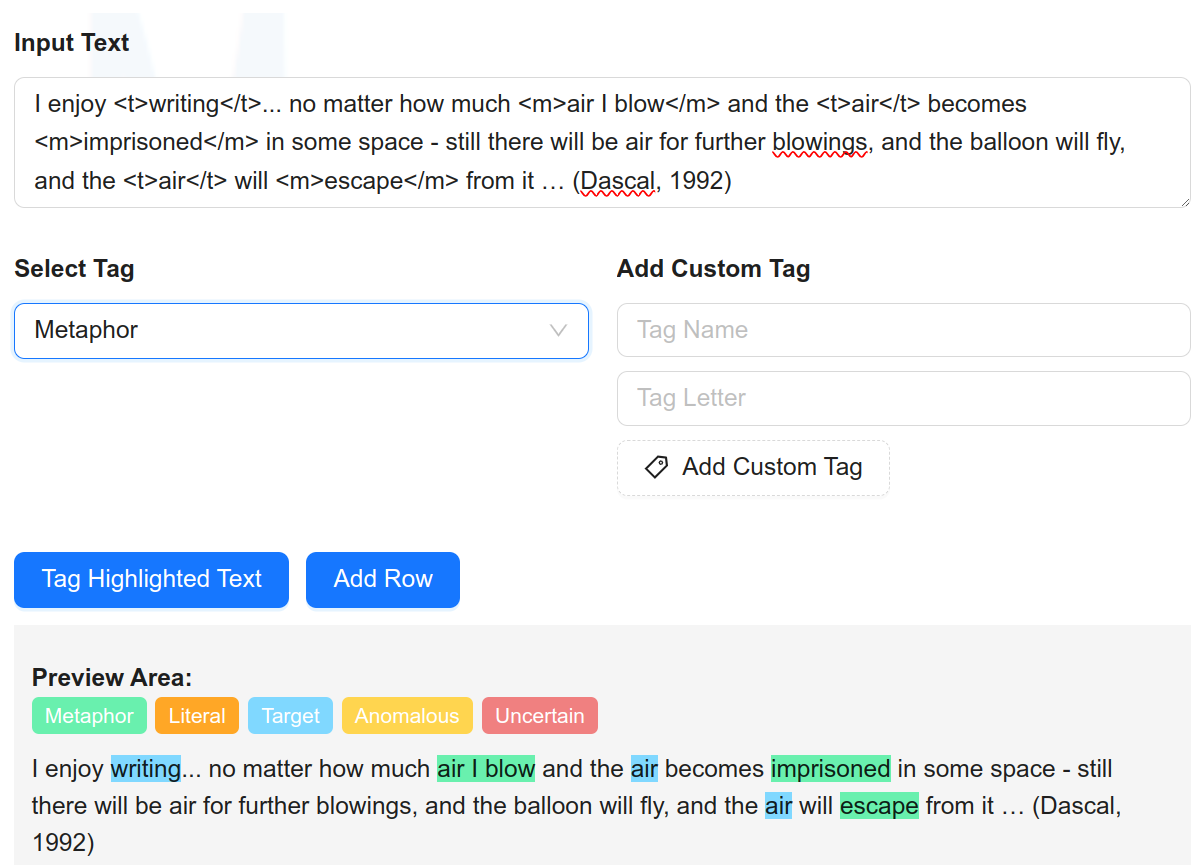}}
\end{center}
\caption{Screenshot of the online annotation tool showing the text input area, tag selection and creation, and resulting tagged text highlighted in different colours.}
\label{fig:annotating-text}
\end{figure*}

\section{Data Sharing} 
\label{sec:formats}

\textsc{MetaphorShare} aims at facilitating computational research on multiple datasets. A minimally constrained format is required to store the datasets into a database, conveniently compare the resources, search efficiently into the different fields across datasets, and run experiments on several sets at once. Metaphor datasets can be added by their authors or by anyone else if their license allows it. In this section, we review the existing dataset formats and our unification choices. The website database currently contains 25 datasets. We select twelve datasets with open licenses and diverse encoded information to illustrate our unification process and as representative examples for our repository. Their characteristics are summarized in Table \ref{tab:ten_datasets} and in more details in the appendix (Table \ref{tab:twelve_datasets}).

\paragraph{One tagged expression per example.} Many datasets in NLP contain labels for binary classification of one delimited expression within a given context. The expressions considered might be single tokens, multi-word expressions (MWE), phrases or compounds. The task of the original data release paper may be to decide weather the marked expression is used metaphorically or literally, or to find a correct literal paraphrase of the metaphoric expression, among other possible analysis and tasks. We define a pair of constrained tags \textit{<m>} and \textit{<l>} to mark metaphoric of literal annotated expressions in text.
Some datasets, e.g. \citet{Jankowiak2020}, also contain anomalous sentences, that are handled with an additional \textit{<a>} tag.

\paragraph{Multiple tags per example.}  \citet{Steen:2010aa} defined an annotation guideline that is widely used in metaphor studies, the MIPVU procedure, together with the release of the VU Amsterdam Corpus (VUAC), the largest existing metaphor corpus. Other examples of English open datasets created following MIPVU, that are in \textsc{MetaphorShare}, are \citet{F04UW5_2019} and \citet{nina-book} on the Music Criticism domain. Many other MIPVU datasets exist in various languages with adaptations of the guideline \cite{:/content/books/9789027261755}, for example \cite{95QQ2W_2022} for Norwegian. In this framework, every token in a document is labelled as metaphoric or literal.
                             
Different versions of the VUAC suiting NLP tasks, adapted for binary classification tasks \cite{leong-etal-2018-report}, or including additional ratings of novelty \cite{Parde_Nielsen_2018} have been published over the years. In the evaluation section, Section \ref{sec:evaluation}, we use a modified version of the VUAC corpus, adapted for binary classification (VUAC\_BO) as described in \citet{boisson-etal-2023-construction}. 
\textsc{MetaphorShare} accepts entries with multiple tags per text span, as exemplified in Figure \ref{fig:annotating-text}. MIPVU datasets are integrated in our repository with two versions fitting different possible NLP tasks: multiple tags per sentence, and one tag per sentence, with duplicated sentences.

\paragraph{Context.} The context of a tagged expression may consist of one single word, in the case of adjective-noun pairs datasets \cite{gutierrez-etal-2016-literal,tsvetkov-etal-2014-metaphor}. It often consist of one sentence such as \citet{mohammad-etal-2016-metaphor} or \citet{Cardillo-2010,Cardillo2017} or \citet{Jankowiak2020}. Longer contexts are sometimes directly provided, such as in \citet{tayyarmadabushi-etal-2022-semeval} or \citet{haagsma-etal-2020-magpie} who include two sentences before and after the sentence containing the marked Potential Idiomatic Expression (PIE), or indirectly provided with a reference to the original document, such as for example in \citet{joseph-etal-2023-newsmet} who share references to newspaper articles or \citet{TuRoth12} who link the annotated examples to sentence pointers in the British National Corpus.We design options to distinguish and display narrow and broad contexts surrounding a tagged expression.

\paragraph{Additional Information.} Beyond metaphorical/literal labels of expressions in context, metaphor datasets may contain
additional information. Some datasets, for example the LCC \cite{mohler-etal-2016-introducing} and \citet{gordon-etal-2015-corpus}, tag the fragment of the context that carries the lexical information about the target of a metaphor (e.g. in the phrase \textit{ocean of happiness}, the metaphoric expression is \textit{ocean} and the target lexical cue of the metaphor is \textit{happiness}). We define a specific tag \textit{<t>} for such lexical cues and a free tag \textit{<u>} for less frequently used in-text annotations. 

Additional information might also be continuous or integer variables such as metaphoricity/figurativeness ratings averages \cite{Cardillo-2010,dunn-2014-measuring, katz-norms}, confidence scores, concreteness scores, novelty scores \cite{Parde_Nielsen_2018}, level of emotion scores \cite{mohammad-etal-2016-metaphor}, frequency in corpora... They may also contain categorical variables such as PoS, textual source and genre, metaphor type, source and target concept/domains information \cite{gordon-etal-2015-corpus}. Figure \ref{fig:search} shows an example of the additional information encoded in the \citet{mohammad-etal-2016-metaphor} dataset for one instance, as diplayed on \textsc{MetaphorShare}. Such features are unified through name recommendations in the guideline.

\paragraph{Unified input format.} 
 
\textsc{MetaphorShare} currently accepts CSV files having minimally one column named \textit{tagged\_text}. It must contain text with at least one tagged expression per line. The accepted tag set is: \textit{< m: metaphoric, l: literal, t: target, a:semantically anomalous,  u: free tag>}. 
Multiple tags per entry are allowed, e.g. \textit{I <m>swim</m> today in an <m>ocean</m> of <t>happiness</t>}. Additional information may be added in freely named columns. Users are encouraged to follow naming recommendations for common features. Sentence indices can be added to preserve the sentence order in a document and handle datasets labeled at the discourse level. All the datasets containing examples in languages other than English could in theory be uploaded in the flexible format defined above, e.g. Polish sentences in \citet{Jankowiak2020}, Mandarin Chinese sentences in \citet{chinese-norms}, Farsi sentences in \citet{levin-etal-2014-resources} and Norwegian examples in \citet{95QQ2W_2022} can be converted smoothly. The free tag \textit{<u>}, open features defined in the data file, and built-in multilingual text search functionalities of the \textsc{MetaphorShare} Elasticsearch database make our framework adaptable to the specificities of a language. Moreover, we will provide individual support. If any author of a dataset encounters issues reformatting a dataset into our expected format, we invite them to contact us by email for support.

\begin{table}[]
    \centering
    \footnotesize
    \begin{tabular}{ll}
    \toprule
        \textbf{ Field}& \textbf{Description} \\\midrule
         \textbf{Required}\\\midrule
         Name& Name of the person uploading data \\
         Email& Address of the person uploading data \\
         Dataset& Name of the dataset \\
         Author& Main author of the dataset \\
         CSV file&Formatted dataset \\
         License&New or existing dataset license\\\midrule
         \textbf{Optional}& \\\midrule
         Paper title&Publication presenting the dataset\\
         Author(s)&Authors of the paper\\
         Year& Paper's publication year\\
         Reference&Bibtex\\
         Languages&Languages of the labelled examples\\\midrule
         Source&Source corpora of the labelled instances\\ 
         Genre&Novel, poetry, news, spoken language...\\\midrule
         Target POS& PoS of the target if any\\
         Source POS& PoS of the source expression if any\\
         Annot. num.& Number of annotators\\
         Annot. profile&Linguists, authors, crowdsourcing...\\
         IAA& Inter-annot. agreement metric \& score\\
         Comments& Additional description and comments\\
    \bottomrule
    \end{tabular}
    \caption{Required and optional dataset information to fill when uploading a new dataset}
    \label{tab:dataset_information}
\end{table}

\section{MetaphorShare: The Website}
\label{sec:metaphorshare}

MetaphorShare general architecture is described in Section \ref{sec:architecture}. The website is organized around four main pages: an uploading page (Section \ref{sec:upload}), a dataset catalog (Section \ref{sec:dataset-catalog}), a page for searching into the datasets records (Section \ref{sec:metaphorshare-search}), and an online annotation tool (Section \ref{sec:online-annotation}).

\subsection{Website architecture}
\label{sec:architecture}

MetaphorShare is hosted by Cardiff University School of Computer Science and Informatics. It employs SSL protocols and HTTPS to safeguard data transmission. Direct access to the database is restricted, with backend schedulers managing all search and upload operations, thus fortifying data protection.

\paragraph{Back \& front end.}
The backend is developed with Python FastAPI, chosen for its speed and ease of use. It features schedulers for routine tasks such as data ingestion and file cleanup. Integration with Elasticsearch allows for efficient indexing and retrieval of dataset entries. Additionally, dataset metadata is securely stored in a PostgreSQL database, ensuring data integrity and structured storage. The frontend is developed using ReactJS (version 18), leveraging libraries such as Bootstrap and Ant Design. State management is handled by Redux, and data visualization is powered by Chart.js.

\paragraph{Database.} Our system employs a dual-database approach to efficiently manage and query data. A PostgreSQL database serves as the primary storage for user-submitted data during the dataset upload process, and stores them ensuring data integrity and facilitating efficient data management.
An Elasticsearch engine enhances the search capabilities of \textsc{MetaphorShare} with exact and fuzzy match for text field supported in multiple languages.

\begin{table}[]
    \centering
    \footnotesize  
    \begin{tabular}{ll}
    \toprule
        \textbf{ Field}& \textbf{Description} \\\midrule
         \textbf{Required}&\\\midrule
          id& Database record ID\\
          dataset\_id&Link to the Dataset index\\
          expression &Token or expression being labelled\\
          label&Metaphoric, literal, anomalous, other\\
          position &Expression offset in the given context\\\midrule
         \textbf{Optional}&(often present)\\\midrule
        sent\_index&Index preserving the sentences order\\
        reference&Reference to a source document\\
        pos& Part of Speech\\
        long\_context&Context surrounding a labelled sentence\\
        target&Lexical cue of the target concept\\
        target\_position&target offset in the given context\\
        source &Metaphor source concept/domain\\
        target &Metaphor target concept/domain\\
        mscore&Metaphoricity/figurativeness score\\\midrule
        \textbf{Free}&Any field, for example \textit{emotion ratings}\\\bottomrule
    \end{tabular}
    \caption{Fields to describe an instance of a dataset in the ElasticSearch database}
    \label{tab:instance-fields}
\end{table}

\subsection{Uploading a new dataset}
\label{sec:upload}

Once a dataset is formatted into a CSV file, as described in Section \ref{sec:formats}, and once its license is specified by the authors, the dataset can be uploaded by any user.


\paragraph{Dataset metadata.} The required and optional fields to be filled in at uploading time are presented in Table \ref{tab:dataset_information}. For example, if a dataset is associated with one or several publications, the references should be added in the appropriate fields, preferably in \textsc{BibTeX} format. The information provided will be displayed in the dataset catalog (c.f. Figure \ref{fig:searchdatasets} in the appendix), and will help to further categorize the datasets in the future versions of the website. 

\paragraph{Automatic and manual validations.} After validating the form, an immediate automatic file format check on the data is done and feedback is returned to the user for acceptance or rejection of the file. Outputs of this step can be seen in Figure \ref{fig:automatic_validation} \& \ref{fig:automatic_validation_accepted} in the appendix. In a second manual validation step, the website administrators check the license information, dataset reference, and eventually suggest modifications of CSV free columns names, that contain the dataset specific additional information, in an email interaction with the user, for database field unification purposes.

\paragraph{Storage in the database.} Once a dataset passes the two validation steps, it is stored in the main repository database. This ElasticSearch database has two indices: the \textit{Datasets} index similar to Table \ref{tab:dataset_information} fields stores all the user provided metadata, and information extracted during the processing of the data file such as the dataset size and the label distribution. The second one, the Potentially Metaphoric Expressions, \textit{PME} index, contains the output of the data file parse, as shown Table \ref{tab:instance-fields}. Each textual labelled example is linked to the \textit{Dataset} index.

\subsection{Dataset Catalog}
\label{sec:dataset-catalog}

\label{sec:metaphorshare-search}

The purpose of this page is to display the main features of each dataset included in the repository. The fields filled at uploading time, including the license chosen, are displayed in the dataset catalog page. Statistics computed from the input file are also shown and plotted, in particular the number of instances, label set and distribution. The list of additional fields is also extracted from the input file. 
A partial view of a dataset summary in the catalog is shown Figure \ref{fig:searchdatasets} in the appendix.

\subsection{Search Page}
\label{sec:metaphorshare-search}

The records can be accessed and downloaded through a search page, as shown in Figure \ref{fig:search}. The search fields include optional selections of datasets and languages from drop-down lists, restricted search for a label (e.g. \textit{metaphorical} or \textit{literal}), and keyword search into the PME text (with exact and fuzzy match), or into the entire text. One phrase or sentence is shown by default on the result page, but longer context can be accessed in one click, as well as all the additional information attached to a record.\footnote{The organization of the catalog and the fine-grained search options are likely to evolve in future versions of \textsc{MetaphorShare}.} The result of any search can be downloaded as a CSV file.

\begin{figure*}[t!]
\includegraphics[width=\textwidth]{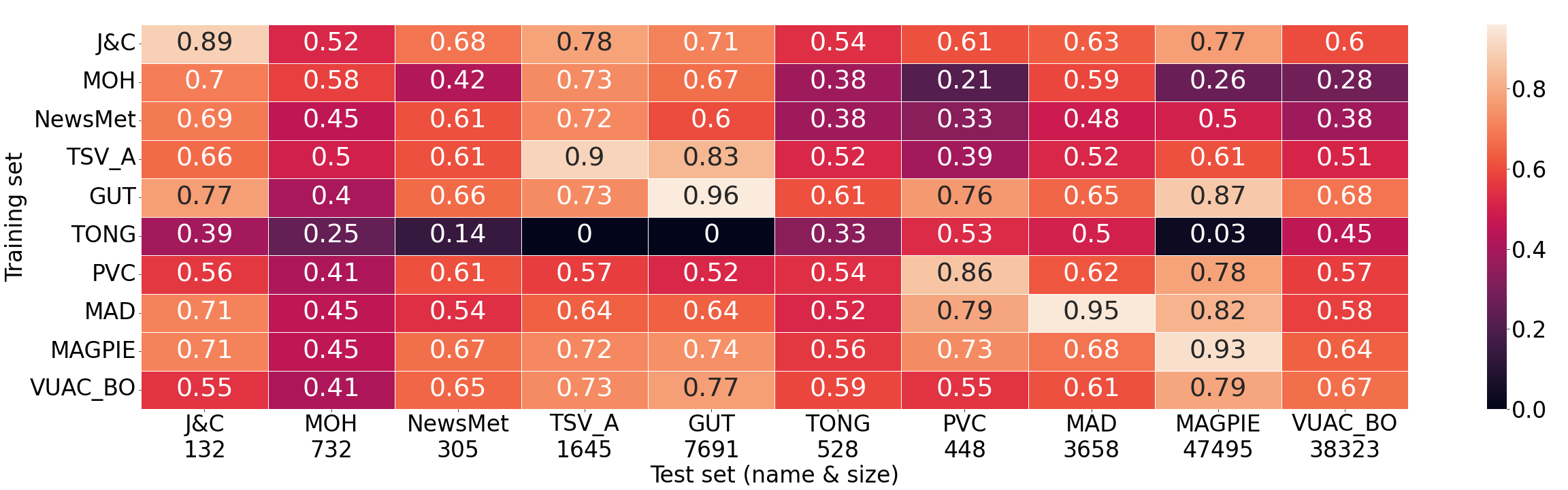}
\caption{Results of the cross dataset evaluation. F1-score of the \textit{metaphor} class. Each training set contains 800 examples and the test sets sizes are shown on the \(x\) axis.}
\label{fig:cross-dataset-results}
\end{figure*}

\subsection{Online data annotation}
\label{sec:online-annotation}
On the Edit/Label Dataset page, one can choose to directly input some text into the interface of the annotation tool, or alternatively to upload a table (in a CSV or Excel format) containing unannotated or partially annotated text. 
An input can then be labelled from scratch or edited by pairing highlighted expressions with tags (c.f. Figure \ref{fig:annotating-text}). While the \textsc{MetaphorShare} tagset is predefined, custom tags can also be created. The resulting labelled data (c.f. Figure \ref{fig:annot-created-records} in the appendix) can then be downloaded as a CSV, or directly uploaded in \textsc{MetaphorShare} following the procedure described in Section \ref{sec:upload}. This online annotation functionality facilitates the creation of resources immediately compatible with our accepted formats, and speeds up the correction of CSV files that do not pass the automatic validation step of the dataset upload procedure. It is also an accessible tool for annotators without any computer science background.

\section{Evaluation}
\label{sec:evaluation}

As a sanity check, we first verified that \textsc{MetaphorShare} uploading, database insertion and search functionalities are working properly with the datasets presented in Table \ref{tab:ten_datasets}. 

Then, as a case study we perform a cross-dataset metaphor identification analysis to illustrate a possible usage of \textsc{MetaphorShare} for NLP research. Given an input expression, the task consists of deciding whether the expression is a metaphor or not. 
Different projects label metaphors for different purposes. \textsc{MetaphorShare} makes it easy to fine-tune models on specific datasets to further support project-specific automatic or semi-automatic labeling.

\paragraph{Experimental setting.} In order to maintain a training set of the same size for all datasets, we randomly sampled 800 examples from each set to create the training data. Because our datasets from psycholinguisics are small and similar in their creation method, we grouped them into one single set called J\&C in this experiment. RoBERTa base models \cite{ROBERTA} are then finetuned independently on the 10 datasets and tested on each of them. Similarly to the experiments in \citet{boisson-etal-2023-construction}, for hyperparameter optimisation, we rely on the Bayesian Optimization with the Hyperband (BOHB) algorithm \cite{falkner2018bohb} available in RayTune \cite{liaw2018tune}, with an identical hyperparameter search space and 25 trials.

\paragraph{Results.} Results are shown in Figure \ref{fig:cross-dataset-results}. As expected, models trained and evaluated on the same dataset often achieved the best results. A few datasets generalise better than others. It is surprisingly the case for J\&C and GUT, both containing only short example with constrained syntactic structures. On the other hand, datasets such as TONG and NEWS do not generalize as well, probably due to different labelled PoS, to the token or MWE span of the labels, or due to redundant contexts corresponding to datasets created for other tasks such as paraphrasing in the case of TONG.

\section{Conclusion and Future Work}

In this paper, we have presented a website to label, unify and share metaphor datasets. It enables an easy integration of new resources of different original formats by the community. There are currently 25 English datasets integrated in the platform. In the future, we are planning to integrate and reach communities working on other languages as one of the main aims.

As far as the platform is concerned, we will focus on integrating a functionality for automatically labelling raw text. The progress recently made by transformer-based language models for metaphor identification opens the path to the creation of helpful personalized tools for automatic or semi-automatic labelling of metaphors based on an initial sample of annotations.

\section*{Acknowledgments}
This research was supported by the Association for Researching and Applying Metaphors through the \textit{Building Bridges} fund. We thank the British National Corpus for giving us the permission to share fragments of texts of the BNC directly on \textsc{MetaphorShare}. We also thank Eileen Cardillo, all the authors of \citet{gutierrez-etal-2016-literal} and of \citet{E06-1042} for allowing us to redistribute their datasets online for research purposes, and to the rest of the authors of the datasets included in the repository for the creation of metaphor datasets with a flexible open license. Without the effort of the metaphor community, this work would not have been possible.
Jose Camacho-Collados was supported by a UKRI Future Leaders Fellowship.

\bibliography{custom,anthology-shrunk}

\begin{thebibliography}{64}
\providecommand{\natexlab}[1]{#1}

\bibitem[{Abzianidze et~al.(2017)Abzianidze, Bjerva, Evang, Haagsma, van Noord, Ludmann, Nguyen, and Bos}]{abzianidze-etal-2017-parallel}
Lasha Abzianidze, Johannes Bjerva, Kilian Evang, Hessel Haagsma, Rik van Noord, Pierre Ludmann, Duc-Duy Nguyen, and Johan Bos. 2017.
\newblock \href {https://aclanthology.org/E17-2039} {The {P}arallel {M}eaning {B}ank: Towards a multilingual corpus of translations annotated with compositional meaning representations}.
\newblock In \emph{Proceedings of the 15th Conference of the {E}uropean Chapter of the Association for Computational Linguistics: Volume 2, Short Papers}, pages 242--247, Valencia, Spain. Association for Computational Linguistics.

\bibitem[{Agerri et~al.(2008)Agerri, Barnden, Lee, and Wallington}]{agerri-etal-2008-textual}
Rodrigo Agerri, John Barnden, Mark Lee, and Alan Wallington. 2008.
\newblock \href {https://aclanthology.org/W08-2228} {Textual entailment as an evaluation framework for metaphor resolution: A proposal}.
\newblock In \emph{Semantics in Text Processing. {STEP} 2008 Conference Proceedings}, pages 357--363. College Publications.

\bibitem[{Baleato~Rodr{\'\i}guez et~al.(2023)Baleato~Rodr{\'\i}guez, Dankers, Nakov, and Shutova}]{baleato-rodriguez-etal-2023-paper}
Daniel Baleato~Rodr{\'\i}guez, Verna Dankers, Preslav Nakov, and Ekaterina Shutova. 2023.
\newblock \href {https://doi.org/10.18653/v1/2023.findings-eacl.35} {Paper bullets: Modeling propaganda with the help of metaphor}.
\newblock In \emph{Findings of the Association for Computational Linguistics: EACL 2023}, pages 472--489, Dubrovnik, Croatia. Association for Computational Linguistics.

\bibitem[{Birke and Sarkar(2006)}]{E06-1042}
Julia Birke and Anoop Sarkar. 2006.
\newblock \href {http://aclweb.org/anthology/E06-1042} {A clustering approach for nearly unsupervised recognition of nonliteral language}.
\newblock In \emph{11th Conference of the European Chapter of the Association for Computational Linguistics}.

\bibitem[{Boisson et~al.(2023)Boisson, Espinosa-Anke, and Camacho-Collados}]{boisson-etal-2023-construction}
Joanne Boisson, Luis Espinosa-Anke, and Jose Camacho-Collados. 2023.
\newblock \href {https://doi.org/10.18653/v1/2023.emnlp-main.406} {Construction artifacts in metaphor identification datasets}.
\newblock In \emph{Proceedings of the 2023 Conference on Empirical Methods in Natural Language Processing}, pages 6581--6590, Singapore. Association for Computational Linguistics.

\bibitem[{Brown et~al.(2020)Brown, Mann, Ryder, Subbiah, Kaplan, Dhariwal, Neelakantan, Shyam, Sastry, Askell, Agarwal, Herbert{-}Voss, Krueger, Henighan, Child, Ramesh, Ziegler, Wu, Winter, Hesse, Chen, Sigler, Litwin, Gray, Chess, Clark, Berner, McCandlish, Radford, Sutskever, and Amodei}]{GPT3}
Tom~B. Brown, Benjamin Mann, Nick Ryder, Melanie Subbiah, Jared Kaplan, Prafulla Dhariwal, Arvind Neelakantan, Pranav Shyam, Girish Sastry, Amanda Askell, Sandhini Agarwal, Ariel Herbert{-}Voss, Gretchen Krueger, Tom Henighan, Rewon Child, Aditya Ramesh, Daniel~M. Ziegler, Jeffrey Wu, Clemens Winter, Christopher Hesse, Mark Chen, Eric Sigler, Mateusz Litwin, Scott Gray, Benjamin Chess, Jack Clark, Christopher Berner, Sam McCandlish, Alec Radford, Ilya Sutskever, and Dario Amodei. 2020.
\newblock Language models are few-shot learners.
\newblock In \emph{Proceedings of the Annual Conference on Neural Information Processing Systems}.

\bibitem[{Cameron and Maslen(2010)}]{cameron2010metaphor}
L.~Cameron and R.~Maslen. 2010.
\newblock \href {https://books.google.fr/books?id=3nL9PgAACAAJ} {\emph{Metaphor Analysis: Research Practice in Applied Linguistics, Social Sciences and the Humanities}}.
\newblock Studies in applied linguistics. Equinox.

\bibitem[{Cardillo et~al.(2017)Cardillo, Watson, and Chatterjee}]{Cardillo2017}
Eileen~R. Cardillo, Christine Watson, and Anjan Chatterjee. 2017.
\newblock \href {https://doi.org/10.3758/s13428-016-0717-1} {Stimulus needs are a moving target: 240 additional matched literal and metaphorical sentences for testing neural hypotheses about metaphor}.
\newblock \emph{Behavior Research Methods}, 49(2):471--483.

\bibitem[{Cardillo et~al.(2010)Cardillo, Schmidt, Kranjec, and Chatterjee}]{Cardillo-2010}
E.R. Cardillo, G.L. Schmidt, A.~Kranjec, and A.~Chatterjee. 2010.
\newblock \href {https://doi.org/10.3758/BRM.42.3.651} {Stimulus design is an obstacle course: 560 matched literal and metaphorical sentences for testing neural hypotheses about metaphor.}

\bibitem[{Chakrabarty et~al.(2021{\natexlab{a}})Chakrabarty, Ghosh, Poliak, and Muresan}]{chakrabarty-etal-2021-figurative}
Tuhin Chakrabarty, Debanjan Ghosh, Adam Poliak, and Smaranda Muresan. 2021{\natexlab{a}}.
\newblock \href {https://doi.org/10.18653/v1/2021.findings-acl.297} {Figurative language in recognizing textual entailment}.
\newblock In \emph{Findings of the Association for Computational Linguistics: ACL-IJCNLP 2021}, pages 3354--3361, Online. Association for Computational Linguistics.

\bibitem[{Chakrabarty et~al.(2021{\natexlab{b}})Chakrabarty, Zhang, Muresan, and Peng}]{chakrabarty-etal-2021-mermaid}
Tuhin Chakrabarty, Xurui Zhang, Smaranda Muresan, and Nanyun Peng. 2021{\natexlab{b}}.
\newblock \href {https://doi.org/10.18653/v1/2021.naacl-main.336} {{MERMAID}: Metaphor generation with symbolism and discriminative decoding}.
\newblock In \emph{Proceedings of the 2021 Conference of the North American Chapter of the Association for Computational Linguistics: Human Language Technologies}, pages 4250--4261, Online. Association for Computational Linguistics.

\bibitem[{Dunn(2014)}]{dunn-2014-measuring}
Jonathan Dunn. 2014.
\newblock \href {https://doi.org/10.3115/v1/P14-2121} {Measuring metaphoricity}.
\newblock In \emph{Proceedings of the 52nd Annual Meeting of the Association for Computational Linguistics (Volume 2: Short Papers)}, pages 745--751, Baltimore, Maryland. Association for Computational Linguistics.

\bibitem[{Falkner et~al.(2018)Falkner, Klein, and Hutter}]{falkner2018bohb}
Stefan Falkner, Aaron Klein, and Frank Hutter. 2018.
\newblock \href {https://arxiv.org/abs/1807.01774} {Bohb: Robust and efficient hyperparameter optimization at scale}.
\newblock \emph{Preprint}, arXiv:1807.01774.

\bibitem[{Fass(1997)}]{Fass1997ProcessingMA}
Dan Fass. 1997.
\newblock Processing metaphor and metonymy.

\bibitem[{Ferraresi et~al.(2008)Ferraresi, Zanchetta, Baroni, and Bernardini}]{Ferraresi2008IntroducingAE}
Adriano Ferraresi, Eros Zanchetta, Marco Baroni, and Silvia Bernardini. 2008.
\newblock \href {https://api.semanticscholar.org/CorpusID:4847295} {Introducing and evaluating ukwac , a very large web-derived corpus of english}.

\bibitem[{Ge et~al.(2023)Ge, Mao, and Cambria}]{Ge2023}
Mengshi Ge, Rui Mao, and Erik Cambria. 2023.
\newblock \href {https://doi.org/10.1007/s10462-023-10564-7} {A survey on computational metaphor processing techniques: From identification, interpretation, generation to application}.
\newblock \emph{Artificial Intelligence Review}, 56(02):1829--1895.

\bibitem[{Ghosh et~al.(2022)Ghosh, Beigman~Klebanov, Muresan, Feldman, Poria, and Chakrabarty}]{flp-2022-figurative}
Debanjan Ghosh, Beata Beigman~Klebanov, Smaranda Muresan, Anna Feldman, Soujanya Poria, and Tuhin Chakrabarty, editors. 2022.
\newblock \href {https://aclanthology.org/2022.flp-1.0} {\emph{Proceedings of the 3rd Workshop on Figurative Language Processing (FLP)}}. Association for Computational Linguistics, Abu Dhabi, United Arab Emirates (Hybrid).

\bibitem[{Ghosh et~al.(2024)Ghosh, Muresan, Feldman, Chakrabarty, and Liu}]{fig-lang-2024-figurative}
Debanjan Ghosh, Smaranda Muresan, Anna Feldman, Tuhin Chakrabarty, and Emmy Liu, editors. 2024.
\newblock \href {https://aclanthology.org/2024.figlang-1.0} {\emph{Proceedings of the 4th Workshop on Figurative Language Processing (FigLang 2024)}}. Association for Computational Linguistics, Mexico City, Mexico (Hybrid).

\bibitem[{Gordon et~al.(2015)Gordon, Hobbs, May, Mohler, Morbini, Rink, Tomlinson, and Wertheim}]{gordon-etal-2015-corpus}
Jonathan Gordon, Jerry Hobbs, Jonathan May, Michael Mohler, Fabrizio Morbini, Bryan Rink, Marc Tomlinson, and Suzanne Wertheim. 2015.
\newblock \href {https://doi.org/10.3115/v1/W15-1407} {A corpus of rich metaphor annotation}.
\newblock In \emph{Proceedings of the Third Workshop on Metaphor in {NLP}}, pages 56--66, Denver, Colorado. Association for Computational Linguistics.

\bibitem[{Guti{\'e}rrez et~al.(2016)Guti{\'e}rrez, Shutova, Marghetis, and Bergen}]{gutierrez-etal-2016-literal}
E.~Dario Guti{\'e}rrez, Ekaterina Shutova, Tyler Marghetis, and Benjamin Bergen. 2016.
\newblock \href {https://doi.org/10.18653/v1/P16-1018} {Literal and metaphorical senses in compositional distributional semantic models}.
\newblock In \emph{Proceedings of the 54th Annual Meeting of the Association for Computational Linguistics (Volume 1: Long Papers)}, pages 183--193, Berlin, Germany. Association for Computational Linguistics.

\bibitem[{Haagsma et~al.(2020)Haagsma, Bos, and Nissim}]{haagsma-etal-2020-magpie}
Hessel Haagsma, Johan Bos, and Malvina Nissim. 2020.
\newblock \href {https://aclanthology.org/2020.lrec-1.35} {{MAGPIE}: A large corpus of potentially idiomatic expressions}.
\newblock In \emph{Proceedings of the Twelfth Language Resources and Evaluation Conference}, pages 279--287, Marseille, France. European Language Resources Association.

\bibitem[{Han et~al.(2022)Han, Mao, and Cambria}]{han-etal-2022-hierarchical}
Sooji Han, Rui Mao, and Erik Cambria. 2022.
\newblock \href {https://aclanthology.org/2022.coling-1.9} {Hierarchical attention network for explainable depression detection on {T}witter aided by metaphor concept mappings}.
\newblock In \emph{Proceedings of the 29th International Conference on Computational Linguistics}, pages 94--104, Gyeongju, Republic of Korea. International Committee on Computational Linguistics.

\bibitem[{Jankowiak(2020)}]{Jankowiak2020}
Katarzyna Jankowiak. 2020.
\newblock \href {https://doi.org/10.1007/s10936-020-09695-7} {Normative data for novel nominal metaphors, novel similes, literal, and anomalous utterances in polish and english}.
\newblock \emph{Journal of Psycholinguistic Research}, 49(4):541--569.

\bibitem[{Joseph et~al.(2023)Joseph, Liu, Ng, See, and Rai}]{joseph-etal-2023-newsmet}
Rohan Joseph, Timothy Liu, Aik~Beng Ng, Simon See, and Sunny Rai. 2023.
\newblock \href {https://doi.org/10.18653/v1/2023.findings-acl.641} {{N}ews{M}et : A {`}do it all{'} dataset of contemporary metaphors in news headlines}.
\newblock In \emph{Findings of the Association for Computational Linguistics: ACL 2023}, pages 10090--10104, Toronto, Canada. Association for Computational Linguistics.

\bibitem[{Julich-Warpakowski(2022)}]{nina-book}
Nina Julich-Warpakowski. 2022.
\newblock \href {https://doi.org/10.1075/milcc.10} {\emph{Motion Metaphors in Music Criticism: An empirical investigation of their conceptual motivation and their metaphoricity}}.
\newblock Metaphor in Language, Cognition, and Communication. John Benjamins Publishing Company.

\bibitem[{Katz et~al.(1988)Katz, Paivio, Marschark, and Clark}]{katz-norms}
Albert Katz, Allan Paivio, Marc Marschark, and Jim Clark. 1988.
\newblock \href {https://doi.org/10.1207/s15327868ms0304_1} {Norms for 204 literary and 260 nonliterary metaphors on 10 psychological dimensions}.
\newblock \emph{Metaphor and Symbol - METAPHOR SYMB}, 3:191--214.

\bibitem[{Kilgarriff et~al.(2014)Kilgarriff, Baisa, Bušta, Jakubíček, Kovář, Michelfeit, Rychlý, and Suchomel}]{kilgarriff2014sketch}
Adam Kilgarriff, Vít Baisa, Jan Bušta, Miloš Jakubíček, Vojtěch Kovář, Jan Michelfeit, Pavel Rychlý, and Vít Suchomel. 2014.
\newblock The sketch engine: ten years on.
\newblock \emph{Lexicography}, 1:7--36.

\bibitem[{Kintsch(2000)}]{Kintsch2000}
Walter Kintsch. 2000.
\newblock \href {https://doi.org/10.3758/BF03212981} {Metaphor comprehension: A computational theory}.
\newblock \emph{Psychonomic Bulletin {\&} Review}, 7(2):257--266.

\bibitem[{Koller et~al.(2008)Koller, Hardie, Rayson, Semino, and Lancaster}]{semantic_anno_2008}
Veronika Koller, Andrew Hardie, Paul Rayson, Elena Semino, and Lancaster. 2008.
\newblock Using a semantic annotation tool for the analysis of metaphor in discourse.
\newblock \emph{Metaphorik.De}, 15.

\bibitem[{Kulkarni et~al.(2024)Kulkarni, Saakyan, Chakrabarty, and Muresan}]{kulkarni-etal-2024-report}
Shreyas Kulkarni, Arkadiy Saakyan, Tuhin Chakrabarty, and Smaranda Muresan. 2024.
\newblock \href {https://aclanthology.org/2024.figlang-1.16} {A report on the {F}ig{L}ang 2024 shared task on multimodal figurative language}.
\newblock In \emph{Proceedings of the 4th Workshop on Figurative Language Processing (FigLang 2024)}, pages 115--119, Mexico City, Mexico (Hybrid). Association for Computational Linguistics.

\bibitem[{Lakoff(1994)}]{lakoff1994master}
G.~Lakoff. 1994.
\newblock \href {https://books.google.fr/books?id=lGSyPgAACAAJ} {\emph{Master Metaphor List}}.
\newblock University of California.

\bibitem[{Lakoff and Johnson(1980)}]{LakoffJohnson80}
George Lakoff and Mark Johnson. 1980.
\newblock \emph{Metaphors we Live by}.
\newblock University of Chicago Press, Chicago.

\bibitem[{Leong et~al.(2018)Leong, Beigman~Klebanov, and Shutova}]{leong-etal-2018-report}
Chee Wee~(Ben) Leong, Beata Beigman~Klebanov, and Ekaterina Shutova. 2018.
\newblock \href {https://doi.org/10.18653/v1/W18-0907} {A report on the 2018 {VUA} metaphor detection shared task}.
\newblock In \emph{Proceedings of the Workshop on Figurative Language Processing}, pages 56--66, New Orleans, Louisiana. Association for Computational Linguistics.

\bibitem[{Li et~al.(2024)Li, Guerin, and Lin}]{li2024findingchallengingmetaphorsconfuse}
Yucheng Li, Frank Guerin, and Chenghua Lin. 2024.
\newblock \href {https://arxiv.org/abs/2401.16012} {Finding challenging metaphors that confuse pretrained language models}.
\newblock \emph{Preprint}, arXiv:2401.16012.

\bibitem[{Liaw et~al.(2018)Liaw, Liang, Nishihara, Moritz, Gonzalez, and Stoica}]{liaw2018tune}
Richard Liaw, Eric Liang, Robert Nishihara, Philipp Moritz, Joseph~E. Gonzalez, and Ion Stoica. 2018.
\newblock \href {https://arxiv.org/abs/1807.05118} {Tune: A research platform for distributed model selection and training}.
\newblock \emph{Preprint}, arXiv:1807.05118.

\bibitem[{Liu et~al.(2019)Liu, Ott, Goyal, Du, Joshi, Chen, Levy, Lewis, Zettlemoyer, and Stoyanov}]{ROBERTA}
Yinhan Liu, Myle Ott, Naman Goyal, Jingfei Du, Mandar Joshi, Danqi Chen, Omer Levy, Mike Lewis, Luke Zettlemoyer, and Veselin Stoyanov. 2019.
\newblock Roberta: A robustly optimized bert pretraining approach.
\newblock \emph{arXiv preprint arXiv:1907.11692}.

\bibitem[{Mao et~al.(2023)Mao, Li, He, Ge, and Cambria}]{mao-etal-2023-metapro}
Rui Mao, Xiao Li, Kai He, Mengshi Ge, and Erik Cambria. 2023.
\newblock \href {https://doi.org/10.18653/v1/2023.acl-demo.12} {{M}eta{P}ro online: A computational metaphor processing online system}.
\newblock In \emph{Proceedings of the 61st Annual Meeting of the Association for Computational Linguistics (Volume 3: System Demonstrations)}, pages 127--135, Toronto, Canada. Association for Computational Linguistics.

\bibitem[{Mao et~al.(2018)Mao, Lin, and Guerin}]{mao-etal-2018-word}
Rui Mao, Chenghua Lin, and Frank Guerin. 2018.
\newblock \href {https://doi.org/10.18653/v1/P18-1113} {Word embedding and {W}ord{N}et based metaphor identification and interpretation}.
\newblock In \emph{Proceedings of the 56th Annual Meeting of the Association for Computational Linguistics (Volume 1: Long Papers)}, pages 1222--1231, Melbourne, Australia. Association for Computational Linguistics.

\bibitem[{Mao et~al.(2019)Mao, Lin, and Guerin}]{mao-etal-2019-end}
Rui Mao, Chenghua Lin, and Frank Guerin. 2019.
\newblock \href {https://doi.org/10.18653/v1/P19-1378} {End-to-end sequential metaphor identification inspired by linguistic theories}.
\newblock In \emph{Proceedings of the 57th Annual Meeting of the Association for Computational Linguistics}, pages 3888--3898, Florence, Italy. Association for Computational Linguistics.

\bibitem[{Martin(1990)}]{Martin:1990}
James~H. Martin. 1990.
\newblock \emph{A Computational Model of Metaphor Interpretation}.
\newblock Academic Press Professional, Inc., San Diego, CA, USA.

\bibitem[{Mohammad et~al.(2016)Mohammad, Shutova, and Turney}]{mohammad-etal-2016-metaphor}
Saif Mohammad, Ekaterina Shutova, and Peter Turney. 2016.
\newblock \href {https://doi.org/10.18653/v1/S16-2003} {Metaphor as a medium for emotion: An empirical study}.
\newblock In \emph{Proceedings of the Fifth Joint Conference on Lexical and Computational Semantics}, pages 23--33, Berlin, Germany. Association for Computational Linguistics.

\bibitem[{Mohler et~al.(2016)Mohler, Brunson, Rink, and Tomlinson}]{mohler-etal-2016-introducing}
Michael Mohler, Mary Brunson, Bryan Rink, and Marc Tomlinson. 2016.
\newblock \href {https://www.aclweb.org/anthology/L16-1668} {Introducing the {LCC} metaphor datasets}.
\newblock In \emph{Proceedings of the Tenth International Conference on Language Resources and Evaluation ({LREC}'16)}, pages 4221--4227, Portoro{\v{z}}, Slovenia. European Language Resources Association (ELRA).

\bibitem[{Nacey(2022)}]{95QQ2W_2022}
Susan Nacey. 2022.
\newblock \href {https://doi.org/10.18710/95QQ2W} {{Replication data for: Systematic metaphors in Norwegian doctoral dissertation acknowledgements}}.

\bibitem[{Nacey et~al.(2019{\natexlab{a}})Nacey, Dorst, Krennmayr, and Reijnierse}]{:/content/books/9789027261755}
Susan Nacey, Aletta~G. Dorst, Tina Krennmayr, and W.~Gudrun Reijnierse, editors. 2019{\natexlab{a}}.
\newblock \href {https://www.jbe-platform.com/content/books/9789027261755} {\emph{Metaphor Identification in Multiple Languages: MIPVU around the world}}.
\newblock John Benjamins.

\bibitem[{Nacey et~al.(2019{\natexlab{b}})Nacey, Krennmayr, Dorst, and Reijnierse}]{F04UW5_2019}
Susan Nacey, Tina Krennmayr, Aletta~G. Dorst, and W.~Gudrun Reijnierse. 2019{\natexlab{b}}.
\newblock \href {https://doi.org/10.18710/F04UW5} {{Replication Data for: What the MIPVU protocol doesn’t tell you (even though it really does)}}.

\bibitem[{Nakayama et~al.(2018)Nakayama, Kubo, Kamura, Taniguchi, and Liang}]{doccano}
Hiroki Nakayama, Takahiro Kubo, Junya Kamura, Yasufumi Taniguchi, and Xu~Liang. 2018.
\newblock \href {https://github.com/doccano/doccano} {{doccano}: Text annotation tool for human}.
\newblock Software available from https://github.com/doccano/doccano.

\bibitem[{Parde and Nielsen(2018)}]{Parde_Nielsen_2018}
Natalie Parde and Rodney Nielsen. 2018.
\newblock \href {https://doi.org/10.1609/aaai.v32i1.11940} {Exploring the terrain of metaphor novelty: A regression-based approach for automatically scoring metaphors}.
\newblock \emph{Proceedings of the AAAI Conference on Artificial Intelligence}, 32(1).

\bibitem[{Pei et~al.(2022)Pei, Ananthasubramaniam, Wang, Zhou, Dedeloudis, Sargent, and Jurgens}]{pei-etal-2022-potato}
Jiaxin Pei, Aparna Ananthasubramaniam, Xingyao Wang, Naitian Zhou, Apostolos Dedeloudis, Jackson Sargent, and David Jurgens. 2022.
\newblock \href {https://doi.org/10.18653/v1/2022.emnlp-demos.33} {{POTATO}: The portable text annotation tool}.
\newblock In \emph{Proceedings of the 2022 Conference on Empirical Methods in Natural Language Processing: System Demonstrations}, pages 327--337, Abu Dhabi, UAE. Association for Computational Linguistics.

\bibitem[{Richards(1936)}]{richards1936philosophy}
I.A. Richards. 1936.
\newblock \href {https://books.google.fr/books?id=AEe-snyOyU4C} {\emph{The Philosophy of Rhetoric}}.
\newblock Bryn Mawr College. Mary Flexner lectures. Oxford University Press.

\bibitem[{Sharma et~al.(2020)Sharma, Bhageria, Scott, PYKL, Das, Chakraborty, Pulabaigari, and Gamb{\"a}ck}]{sharma-etal-2020-semeval}
Chhavi Sharma, Deepesh Bhageria, William Scott, Srinivas PYKL, Amitava Das, Tanmoy Chakraborty, Viswanath Pulabaigari, and Bj{\"o}rn Gamb{\"a}ck. 2020.
\newblock \href {https://doi.org/10.18653/v1/2020.semeval-1.99} {{S}em{E}val-2020 task 8: Memotion analysis- the visuo-lingual metaphor!}
\newblock In \emph{Proceedings of the Fourteenth Workshop on Semantic Evaluation}, pages 759--773, Barcelona (online). International Committee for Computational Linguistics.

\bibitem[{Steen(2010)}]{Steen:2010aa}
Gerard Steen. 2010.
\newblock \emph{A method for linguistic metaphor identification: from MIP to MIPVU}, volume v. 14 of \emph{Converging evidence in language and communication research}.
\newblock John Benjamins Pub. Co., Amsterdam.

\bibitem[{Stowe et~al.(2021)Stowe, Chakrabarty, Peng, Muresan, and Gurevych}]{stowe-etal-2021-metaphor}
Kevin Stowe, Tuhin Chakrabarty, Nanyun Peng, Smaranda Muresan, and Iryna Gurevych. 2021.
\newblock \href {https://doi.org/10.18653/v1/2021.acl-long.524} {Metaphor generation with conceptual mappings}.
\newblock In \emph{Proceedings of the 59th Annual Meeting of the Association for Computational Linguistics and the 11th International Joint Conference on Natural Language Processing (Volume 1: Long Papers)}, pages 6724--6736, Online. Association for Computational Linguistics.

\bibitem[{Stowe et~al.(2022)Stowe, Utama, and Gurevych}]{stowe-etal-2022-impli}
Kevin Stowe, Prasetya Utama, and Iryna Gurevych. 2022.
\newblock \href {https://doi.org/10.18653/v1/2022.acl-long.369} {{IMPLI}: Investigating {NLI} models{'} performance on figurative language}.
\newblock In \emph{Proceedings of the 60th Annual Meeting of the Association for Computational Linguistics (Volume 1: Long Papers)}, pages 5375--5388, Dublin, Ireland. Association for Computational Linguistics.

\bibitem[{Tayyar~Madabushi et~al.(2022{\natexlab{a}})Tayyar~Madabushi, Gow-Smith, Garcia, Scarton, Idiart, and Villavicencio}]{tayyar-madabushi-etal-2022-semeval}
Harish Tayyar~Madabushi, Edward Gow-Smith, Marcos Garcia, Carolina Scarton, Marco Idiart, and Aline Villavicencio. 2022{\natexlab{a}}.
\newblock \href {https://doi.org/10.18653/v1/2022.semeval-1.13} {{S}em{E}val-2022 task 2: Multilingual idiomaticity detection and sentence embedding}.
\newblock In \emph{Proceedings of the 16th International Workshop on Semantic Evaluation (SemEval-2022)}, pages 107--121, Seattle, United States. Association for Computational Linguistics.

\bibitem[{Tayyar~Madabushi et~al.(2022{\natexlab{b}})Tayyar~Madabushi, Gow-Smith, Garcia, Scarton, Idiart, and Villavicencio}]{tayyarmadabushi-etal-2022-semeval}
Harish Tayyar~Madabushi, Edward Gow-Smith, Marcos Garcia, Carolina Scarton, Marco Idiart, and Aline Villavicencio. 2022{\natexlab{b}}.
\newblock {SemEval-2022 Task 2}: {Multilingual Idiomaticity Detection and Sentence Embedding}.
\newblock In \emph{Proceedings of the 16th International Workshop on Semantic Evaluation (SemEval-2022)}. Association for Computational Linguistics.

\bibitem[{Tong et~al.(2024)Tong, Choenni, Lewis, and Shutova}]{tong2024metaphor}
Xiaoyu Tong, Rochelle Choenni, Martha Lewis, and Ekaterina Shutova. 2024.
\newblock \href {https://arxiv.org/abs/2403.11810} {Metaphor understanding challenge dataset for llms}.
\newblock \emph{Preprint}, arXiv:2403.11810.

\bibitem[{Tsvetkov et~al.(2014)Tsvetkov, Boytsov, Gershman, Nyberg, and Dyer}]{tsvetkov-etal-2014-metaphor}
Yulia Tsvetkov, Leonid Boytsov, Anatole Gershman, Eric Nyberg, and Chris Dyer. 2014.
\newblock \href {https://doi.org/10.3115/v1/P14-1024} {Metaphor detection with cross-lingual model transfer}.
\newblock In \emph{Proceedings of the 52nd Annual Meeting of the Association for Computational Linguistics (Volume 1: Long Papers)}, pages 248--258, Baltimore, Maryland. Association for Computational Linguistics.

\bibitem[{Tu and Roth(2012)}]{TuRoth12}
Yuancheng Tu and Dan Roth. 2012.
\newblock Sorting out the most confusing english phrasal verbs.
\newblock In \emph{Proceedings of the First Joint Conference on Lexical and Computational Semantics}.

\bibitem[{Turney et~al.(2011)Turney, Neuman, Assaf, and Cohen}]{turney-etal-2011-literal}
Peter Turney, Yair Neuman, Dan Assaf, and Yohai Cohen. 2011.
\newblock \href {https://aclanthology.org/D11-1063} {Literal and metaphorical sense identification through concrete and abstract context}.
\newblock In \emph{Proceedings of the 2011 Conference on Empirical Methods in Natural Language Processing}, pages 680--690, Edinburgh, Scotland, UK. Association for Computational Linguistics.

\bibitem[{Veale(2016)}]{veale-2016-round}
Tony Veale. 2016.
\newblock \href {https://doi.org/10.18653/v1/W16-1105} {Round up the usual suspects: Knowledge-based metaphor generation}.
\newblock In \emph{Proceedings of the Fourth Workshop on Metaphor in {NLP}}, pages 34--41, San Diego, California. Association for Computational Linguistics.

\bibitem[{Veale and Li(2012)}]{veale-li-2012-specifying}
Tony Veale and Guofu Li. 2012.
\newblock \href {https://aclanthology.org/P12-3002} {Specifying viewpoint and information need with affective metaphors: A system demonstration of the metaphor-magnet web app/service}.
\newblock In \emph{Proceedings of the {ACL} 2012 System Demonstrations}, pages 7--12, Jeju Island, Korea. Association for Computational Linguistics.

\bibitem[{Wachowiak and Gromann(2023)}]{wachowiak-gromann-2023-gpt}
Lennart Wachowiak and Dagmar Gromann. 2023.
\newblock \href {https://doi.org/10.18653/v1/2023.acl-long.58} {Does {GPT}-3 grasp metaphors? identifying metaphor mappings with generative language models}.
\newblock In \emph{Proceedings of the 61st Annual Meeting of the Association for Computational Linguistics (Volume 1: Long Papers)}, pages 1018--1032, Toronto, Canada. Association for Computational Linguistics.

\bibitem[{Wilks(1973)}]{wilks-pref}
Yorick Wilks. 1973.
\newblock \href {https://apps.dtic.mil/sti/citations/AD0764652} {Preference semantics}.

\bibitem[{Zeng and Bhat(2021)}]{zeng-bhat-2021-idiomatic}
Ziheng Zeng and Suma Bhat. 2021.
\newblock \href {https://doi.org/10.1162/tacl_a_00442} {Idiomatic expression identification using semantic compatibility}.
\newblock \emph{Transactions of the Association for Computational Linguistics}, 9:1546--1562.

\end{thebibliography}
\bibliographystyle{acl_natbib}

\appendix

\section{Detailed information of datasets used in our experiments }

Table \ref{tab:twelve_datasets} provides detailed information about the datasets, including license, source corpora, label and sentence distribution. Additional links and preprocessing details are listed below :

\begin{description}

\item[JANK:] Anomalous sentences and simile are not shown in the table.
because because they are not used in our binary classification evaluation task.

\item[MOH:] The license is available at \url{https://saifmohammad.com/WebPages/metaphor.html}. The original dataset contains 1639 instances. 
A few duplicated example sentences caused by orthographic variants of the target word, such as distil/distill have been removed.

\item[NewsMet:] We show the manually labelled sentences (named \textit{gold} by the authors). The corpus from which the sentences are extracted can be found at \url{https://github.com/several27/FakeNewsCorpus}.
\item[TSVET\_A:] The license can be found at \url{https://github.com/ytsvetko/metaphor/blob/master/LICENSE.md}

\item[GUT:] The UKaC is presented in \citet{Ferraresi2008IntroducingAE}.

\item[PVC:] The original dataset is accessible at \url{https://cogcomp.seas.upenn.edu/page/resource_view/26}

\item[MAGPIE:] The PMB corpus from which some sentences are sourced is presented in \citet{abzianidze-etal-2017-parallel}.

\item[TONG:] Original VUAC sentences and \textit{apt} (dataset new label) paraphrases are counted in the table.

\end{description}

\section{Dataset upload}
\label{sec:appendix-upload}

\begin{figure*}[t!]
\efbox{\includegraphics[width=.985\textwidth]{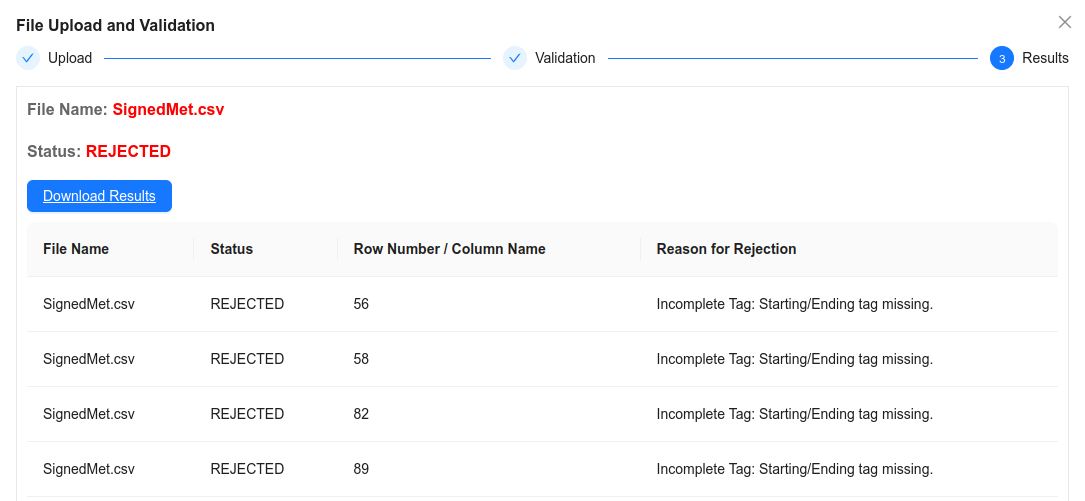}}
\caption{Screenshot of the file format check for a rejected file. The line the error occurs in the CSV file and the type of errors are specified.}
\label{fig:automatic_validation}
\end{figure*}

\begin{figure*}[t!]
\efbox{
\includegraphics[width=.98\textwidth]{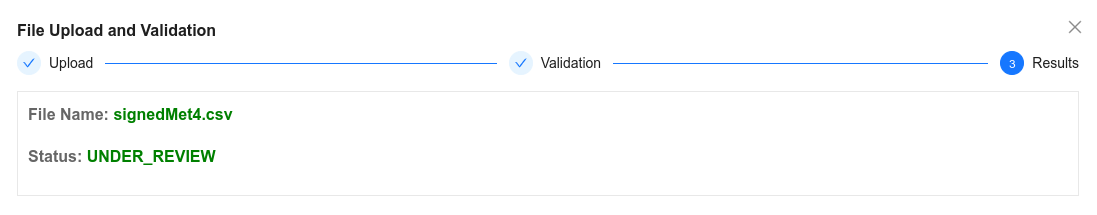}}
\caption{Screenshot of the file format check after a CSV file is accepted for manual review.}
\label{fig:automatic_validation_accepted}
\end{figure*}

\paragraph{Automatic file format check.} Figure \ref{fig:automatic_validation} \& \ref{fig:automatic_validation_accepted} show the row by row feedback provided after the automatic validation step when the file is rejected, and the pop up message indicating that a file passed the automatic validation step.
\paragraph{Dataset information after validation.}
\label{sec:appendix-dataset-information}
Figure \ref{fig:searchdatasets} shows a part of a dataset presentation page once it is integrated into the catalog and database.


\section{Dataset catalog}
\label{appendix:catalog}
\begin{figure*}[]
\efbox{\includegraphics[width=.985\textwidth, height=11cm]{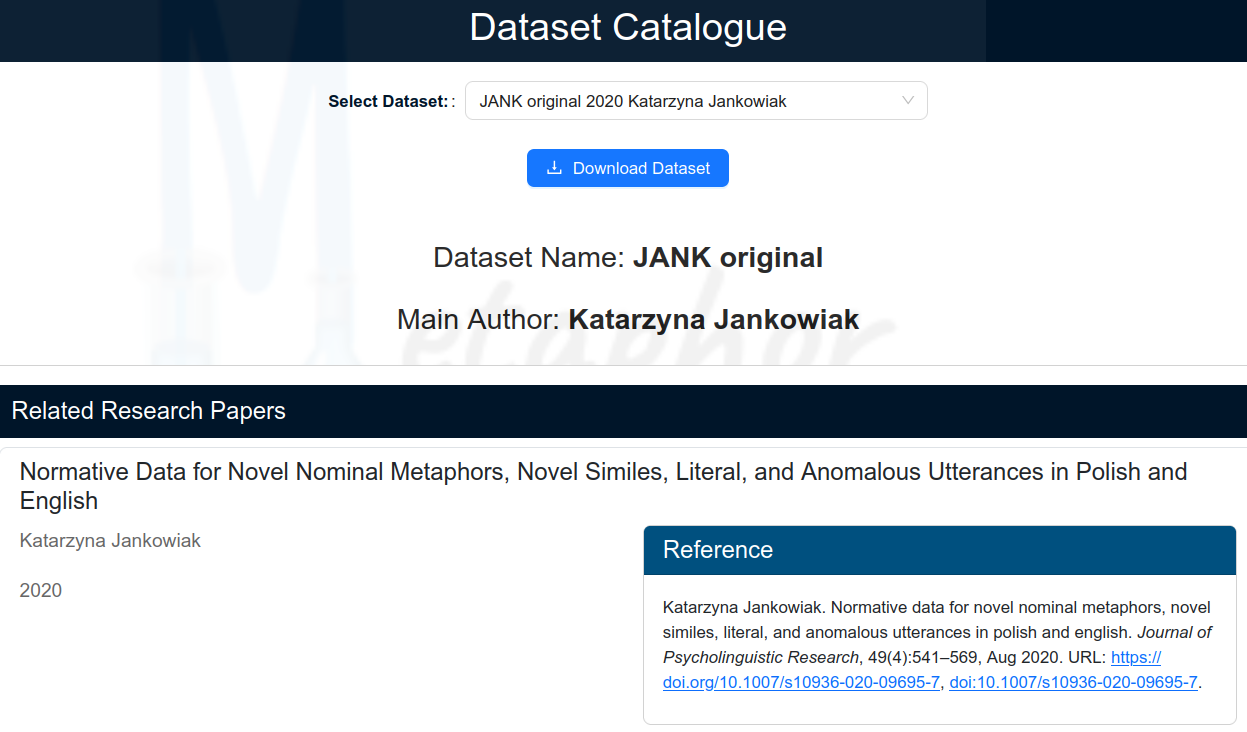}}
\caption{Screenshot of top of the the datasets information page in the catalog section of the website. The English dataset released with \citet{Jankowiak2020} is presented as an example.} 
\label{fig:searchdatasets}
\end{figure*}

Figure \ref{fig:searchdatasets} is a screenshot of the top of the page presenting a dataset in the Catalog section of \textsc{MetaphorShare}. The dataset name, main author and research paper associated to the data release are displayed.

\section{Online annotation tool}
 
\begin{figure*}[t!]
\efbox{
\includegraphics[width=.98\textwidth]{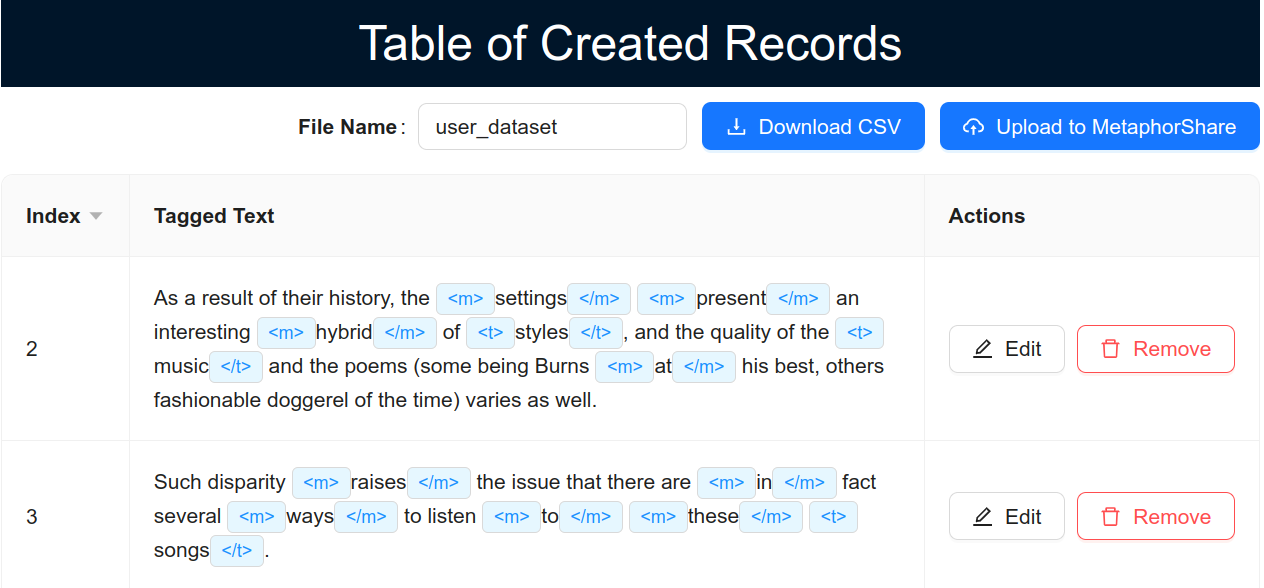}}
\caption{Screenshot showing dataset rows available for tagging or edition, as displayed in the annotation tool.}
\label{fig:annot-created-records}
\end{figure*}

Figure \ref{fig:annot-created-records} shows how the created records of a dataset are displayed, allowing an easy step by step annotation.

\begin{table*}[]
    \centering
    \footnotesize
    \begin{tabular}{l@{\hspace{2pt}}l@{\hspace{2pt}}l@{\hspace{2pt}}l@{\hspace{2pt}}l@{\hspace{2pt}}l@{\hspace{2pt}}l@{\hspace{6pt}}l@{\hspace{6pt}}l}
    \toprule
    \textbf{Dataset}&
    \textbf{Reference} &
    \textbf{License}&
    \textbf{N}&
    \textbf{N dist.}&
    \textbf{N dist.} &
    \textbf{\%}&
    \textbf{Expr.}&
    \textbf{Domain/Source}\\
         \textbf{Name}&\textbf{} &\textbf{}&&\textbf{ctxt.}&\textbf{expr.} &\textbf{met.}&\textbf{PoS}&\\\midrule
         \multicolumn{5}{l}{\textbf{Words in syntactically constrained sentences (Psycholinguistics)}}  &&&&\\\midrule
         
         JANK&\citet{Jankowiak2020}&CC BY 4.0&240&240&120&50&N&constructed examples\\
         CARD\_V&\citet{Cardillo-2010}&CC BY-NC&280&280&140&50&V&constructed examples\\
          CARD\_N&\citet{Cardillo-2010,Cardillo2017}&CC BY-NC&512&512&256&50&N&constructed examples\\\midrule
          \multicolumn{5}{l}{
          \textbf{Words in natural short contexts (NLP)}}&&&&\\\midrule
         MOH&\citet{mohammad-etal-2016-metaphor}&see data page&1632&1632&439&25&V&WordNet examples\\
        NewsMet&\citet{joseph-etal-2023-newsmet}&Apache-2.0&1205&1205&477&49&V&Fake News Corpus\\
     TSV\_A & \citet{tsvetkov-etal-2014-metaphor}&see data page&1945&1072&687&50&A&various websites\\
        GUT&\citet{gutierrez-etal-2016-literal}  &AFL-3.0&8591&3479&23&54&A&Wikipedia, UKWaC...\\\midrule
        \multicolumn{5}{l}{\textbf{Multi-Word Expressions in natural long contexts (NLP)}}  &&&&\\\midrule
        PVC& \citet{TuRoth12}&no license&1348&1348&23&65&V-Prep&BNC\\
        MAD& \citet{tayyarmadabushi-etal-2022-semeval}&GPL-3.0&4558&4554&251&48&NC&Common Crawl\\
        MAGPIE& \citet{haagsma-etal-2020-magpie}&CC-BY-4.0&48395&47283& 9307&75&Various&BNC, PMB. \\\midrule 
        \multicolumn{5}{l}{
        \textbf{Words sampled from VUAC (MIPVU)}}&&&&\\\midrule
        VUAC\_BO  &\citet{boisson-etal-2023-construction} &CC BY-SA 3.0&39223&11476&8674&52&Various&VUAC\\
        TONG&\citet{tong2024metaphor}&CC-BY-4.0&1428&739&861&46&Various&VUAC \& paraphrases\\
    \bottomrule
    \end{tabular}
    \caption{Twelve example metaphor datasets included in the \textsc{MetaphorShare} repository. The N column shows the number of instances in the datasets, followed by the number of distinct provided context surrounding the potential metaphoric expression (N dist ctxt), and the number of distinct lemmas labelled metaphoric or literal (N dist. expr.). \%met. indicates the percentage of expressions labelled as metaphorical, and Expr. PoS their part of speech.}
    \label{tab:twelve_datasets}
\end{table*}

\end{document}